\documentclass[opinion]{sigirforum}

\def\pubissue{Vol. 58 No. 1 -- June 2024}

\usepackage{booktabs}
\usepackage{url}
\usepackage{xspace}
\usepackage{textcomp}
\usepackage{amsmath}
\usepackage{algorithm}
\usepackage[noend]{algpseudocode}
\usepackage{xspace}
\usepackage{balance}
\usepackage{array}
\usepackage{caption}
\usepackage{color}
\usepackage{subcaption}
\usepackage{multirow}
\usepackage{comment}
\usepackage{hyperref}
\usepackage{url}
\usepackage{makecell}
\usepackage{lineno}

\begin{document}
\title{Bridging the Gap Between Information Seeking and Product Search Systems: Q\&A Recommendation for E-commerce}

\authors{
\author[skuzi@amazon.com]{Saar Kuzi}{Amazon.com, Inc.}{Seattle, WA, USA}
\and
\author[malmasi@amazon.com]{Shervin Malmasi}{Amazon.com Inc.}{Seattle, WA, USA}
}

\maketitle 
\begin{abstract}
Consumers on a shopping mission often leverage both product search and information seeking systems, such as web search engines and Question Answering (QA) systems, in an iterative process to improve their understanding of available products and reach a purchase decision. While product search is useful for shoppers to find the actual products meeting their requirements in the catalog, information seeking systems can be utilized to answer any questions they may have to refine those requirements. The recent success of Large Language Models (LLMs) has opened up an opportunity to bridge the gap between the two tasks to help customers achieve their goals quickly and effectively by integrating conversational QA within product search. In this paper, we propose to recommend users Question-Answer (Q\&A) pairs that are relevant to their product search and can help them make a purchase decision. We discuss the different aspects of the problem including the requirements and characteristics of the Q\&A pairs, their generation, and the optimization of the Q\&A recommendation task. We highlight the challenges, open problems, and suggested solutions to encourage future research in this emerging area.
\end{abstract}

\section{Introduction}
\label{sec:intro}

Currently, most shoppers looking to purchase a product online rely on both e-commerce and information seeking systems. Search engines in online shopping websites are built on top of a catalog of the available products and are used by shoppers to explore the different offerings, obtain product specific information, and finally make a purchase \citep{sondhi2018taxonomy}. 
Information seeking systems (mainly web search engines, but more recently also question answering systems) are used by shoppers to complement their product search. This is because users often do not have enough knowledge to confidently make a purchase decision and thus leverage information seeking systems to acquire relevant knowledge to assist them in their shopping journey \citep{rowley2000product,branco2012optimal}. 

\begin{figure}[t]
\centering
\includegraphics[width=1\textwidth]{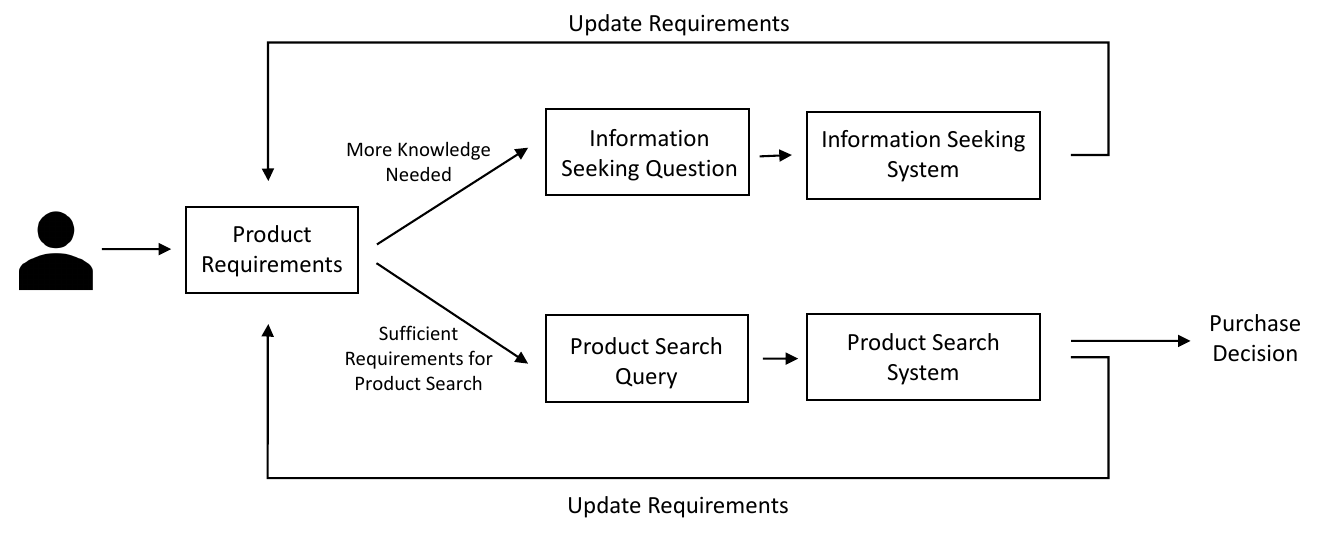}
    \caption{The current online shopping process. Users are switching between information seeking systems and product search to reach a purchase decision. The dashed arrow represents our envisioned approach to bridge the gap between the two systems.}
    \label{fig:example}
\end{figure}

This process, however, is not optimal. First, users that search for products may not have enough knowledge to ask the right questions in an information seeking system. Second, an independent information seeking system does not directly take into account the full context of the user in product search (e.g., the product detail page that the user is viewing or the set of search results).  For those reasons, as illustrated in Figure \ref{fig:example}, users would often need to frequently switch between the two systems to refine their requirements and finally reach a purchase decision. This results in a lengthy process requiring substantial user effort to achieve their shopping goals. Some recent work has attempted to address these issues by enabling product search systems to answer informational queries \citep{chen2023industry}, but combining these distinct features in a single interface may not be intuitive for users.

\begin{figure}[t!]
\centering
\begin{tabular}{c}
\includegraphics[width=7cm,height=8cm]{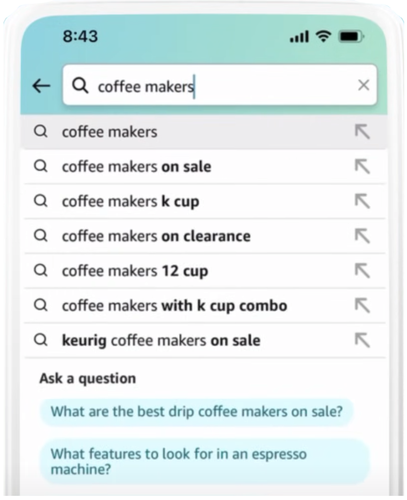}
 \\
\end{tabular}
\caption{\label{fig:auto} An illustration of the suggested questions interface for the auto-complete stage.}
\end{figure}

In this opinion paper, we propose a new product search paradigm to bridge the gap between information seeking and product search systems. Specifically, the idea is to suggest Q\&A pairs to shoppers at different places in a product search system, as illustrated in Figure \ref{fig:auto}. This can assist them in acquiring important purchase-related information with minimal effort. This new direction is quickly becoming a reality, as exemplified by Amazon's new Rufus conversational shopping experience launched in February 2024.\footnote{\url{https://www.aboutamazon.com/news/retail/amazon-rufus}} We discuss the different aspects of the proposed approach including the characteristics of the suggested questions and answers, Q\&A generation approaches, quality control, and the optimization of the recommendation task. The paper can serve as a roadmap for implementing such a system, and highlights the different challenges and open problems that need to be addressed.

The rest of the paper is organized as follows. We begin by explaining the shopping journey of users in online shopping and how it intersects with the suggested Q\&As. Then, we move to discuss the different requirements from the questions and answers in such a system, generation approaches, and quality control. We conclude the paper with a discussion about different directions for addressing the optimization of the recommendation system with user feedback.

\section{Q\&A and the Shopping Journey}
To implement a useful Q\&A suggestion system in product search, it is crucial to understand the \textit{shopping journey} of online customers and how Q\&As can be integrated in it successfully. Online shopping users start their journey with a shopping goal which is a set of requirements from a desired product. The users would then go through several stages that dictate the specificity of those requirements. In general, users would come with an initial set of requirement that are expected to narrow down toward the chosen product as they view different options from the catalog and acquire any necessary knowledge related to their search. Users in product search would usually be in one of the following stages: 
\begin{enumerate}
\item Exploration: users are at the beginning of their shopping journey and are learning about the product space and are refining their requirements. 
\item Comparison: users have narrowed down their requirements and are comparing across several products that meet them.
\item Final Consideration: users are focused on a small set of products and ask specific questions about them to make a purchase decision.
\end{enumerate}

\begin{table}[]
\centering
\caption{\label{tab:examples} Examples of suggested questions in different entry points of a product search system.}
\begin{tabular}{cl}
\toprule
\multicolumn{2}{c}{\textbf{Auto-complete}} \\ \midrule
Context & ``coffee'' \\ \hline
\multirow{3}{*}{Questions} & What are the different types of coffee machines? \\ 
& How to choose a coffee table for your space? \\ 
& What coffee mugs are in style? \\ \hline

Context & ``juicer'' \\ \hline
\multirow{3}{*}{Questions} & What features should I look for when buying a juicer? \\ 
& What are the best juicers for beginners? \\ 
& What types of juicers are good for leafy greens? \\

\hline\midrule

\multicolumn{2}{c}{\textbf{Search Result Page}} \\ \midrule
Context & ``coffee machine single cup''\\ \hline
\multirow{3}{*}{Questions} & How do single cup coffee machines work? \\
& What maintenance is required for single-cup coffee machines? \\
& What are the differences between single-cup machines and traditional ones? \\ \hline

Context & ``cold press juicer''\\ \hline
\multirow{3}{*}{Questions} & How do cold press juicers compare to centrifugal juicers? \\
& What are the pros and cons of vertical vs. horizontal cold press juicers?\\
& What are some popular brands of cold press juicers?\\

\hline\midrule

\multicolumn{2}{c}{\textbf{Product Detail Page}}\\ \midrule
Context & ``Amazon Basics Single Cup Coffee Machine'' \\ \hline
\multirow{3}{*}{Questions} & Is it easy to clean? \\
& What are the dimensions? \\
& Is the tray adjustable? \\ \hline

Context & ``Vertical Cold Press Juicer for Fruits and Vegetables'' \\ \hline
\multirow{3}{*}{Questions} & What do customers think about its durability? \\
& Is it noisy? \\
& Are the parts dishwasher safe?\\

\bottomrule

\end{tabular}
 
\end{table}

From the system perspective, the different stages correspond to different user experiences in the product search system which result in different modalities of Q\&A recommendations, as illustrated in Table \ref{tab:examples}. The first one is \textbf{auto-complete} Q\&A suggestion. Specifically, the idea is to present broad and exploratory related questions while users type their product search query (see Figure~\ref{fig:auto}). The second one is \textbf{Search Result Page (SERP)} Q\&A suggestion. The idea here is to suggest questions that are embedded in the SERP. This is similar to the ``People Also Asked'' suggested questions in web search engines \citep{matias2017generating} but would need to include shopping specific questions that are related to the products being shown. The final modality is \textbf{product detail page} Q\&A suggestion. The idea here is to suggest Q\&A pairs that are relevant to a specific product. 

The different entry points, since users encounter them in different stages of their shopping journey, require different types of Q\&As. For example, questions in the SERP are expected to be broader than those in the product detail page since users have not yet narrowed down their requirements to a specific product and are still learning about the different options. Similarly, questions in the auto-complete modality are expected to be even broader than SERP questions because the shopping goal of the user at this stage is unclear to the system.

\section{Q\&A Requirements}
\label{sec:questions}

Presenting users with the most appropriate questions to their shopping goal and the specific stage of their shopping journey is crucial for the success of the Q\&A system. First, we note that questions can represent various \textit{intents} that can be relevant to different stages in the shopping journey. For example, a question which asks to compare between two products or product categories (e.g., ``What is the difference between single-cup and traditional coffee machines?'') can be appropriate for the exploration and comparison stages but not as much to the detail page of a product. On the other hand, a question with the intent of learning about an aspect of a product (e.g., ``What is the voltage of this coffee machine?'') is more appropriate once the user narrowed down their search to very few options. Some of the popular intents include the following ones.

\begin{itemize}
\item Aspect: Seeking information about an aspect of a specific product or a group of products. 
\item Comparison: Comparing two or more products, or different aspects of the same product.
\item General Knowledge: Seeking some general knowledge that may help the user choose the right product. 
\item Superlative: Asking for the best products on a given aspect. 
\item How-to: Seeking guidance on how to use a specific product or a group of products.
\item Offer: Various topics like pricing, shipping, return policy, refund policy, product availability, and warranty.
\item Subjective: Seeking opinions of other users.
\item Search: Recommendation for products that meet some requirements.
\end{itemize}

Across the different intents, there can be several common requirements for the questions in a specific shopping journey stage. A minimal requirement is for the questions to be \textit{relevant} to the context (e.g., the query or the product detail page). The question must also be \textit{useful} for making a purchase decision given the current stage of the user. Another important requirement is \textit{specificity}. We want to a question to as specific as needed given the user background and the shopping journey phase. Finally, we want the questions to be \textit{concise} by including only the necessary information to users to accommodate small screens and the short attention span of users.

As for the answers to the questions, there are also several general requirements that we need to consider in the context of product search. The minimal requirement from the answer is to be factually correct. This is a very important requirement as users are relying on those recommendations when buying a product and may lose their trust in the system if wrong information led them to make a wrong purchase decision. Since the answers are within a product search system, another requirement is for them to be always shopping related to help users in their shopping journey. For example, if the question is ``What is the difference between espresso and drip coffee?'', the answer may also recommend related products such as espresso machines. Another important consideration is for the answer to leverage the right type of information based on the question intent. For instance, while subjective questions would mainly rely on customer reviews, product aspect questions would leverage the product catalog. Finally, similarly to questions, we would like the answers to be informative but concise.

\section{Q\&A Generation}
\label{sec:answers}

To obtain the questions and answers for Q\&A recommendation, one common approach used in the previous work relies on query logs to mine popular user questions \citep{mitra2021zero,mitra2020transformer}. Then, based on the questions, answers can be generated using abstractive QA from relevant documents. This approach has several limitations in the context of product search engines. First, users in product search engines mainly use keyword queries and do not write fully formed questions, making approaches that rely on logs not applicable. Second, product search engines are built on top of product catalogs and do not typically have easy access to external knowledge. 

For those reasons, we propose to use LLMs for the generation of Q\&A pairs. Using LLMs is advantageous for the task since they can generate both the question and the answer from their internal knowledge \citep{hamalainen2023evaluating,hartvigsen2022toxigen}. To generate questions that are useful for the context, we can provide the context information as part of the input (e.g., the query, items in the SERP, or the product detail page). LLMs also have good capability to process long inputs, which means that we can provide it with different types of data to generate a variety of interesting questions. For example, for product detail page questions we can use customer reviews, the product specifications, and relevant external knowledge to generate questions. Some of these ideas have been explored by \citet{vedula2024question} in the context of shopping assistants. 

Using an LLM, we are also interested in controlling the different aspects of a question to satisfy the different requirements such as specificity and intent. While this can be potentially achieved through appropriate prompting, there can be a limit on the instruction following capability of the model given a very long prompt. For that reason, one option can be to create different prompts for the various scenarios to optimize the question generation for them. For example, we expect the instructions for question generation for SERP to be different than the ones of the product page. Finally, it is also important to note that some aspects that are needed to be controlled, such as intent, are domain specific, making it potentially necessary to further fine-tune the LLM to follow the instructions successfully.

For answer generation, similarly to the case of questions, a promising technique would be to use retrieval augmented generation with an LLM. One key consideration in doing that is to decide which retrieval evidences are relevant to the specific question. For example, while customer reviews are the most useful for subjective questions, product aspect questions may benefit the most from information in the product catalog. 
Ideally, we can provide all of the different sources of retrieval to the LLM. This approach, however, may have some disadvantages such as input length limit, latency, and sensitivity to noise. An alternative approach can be to leverage different retrieval sources for different types of questions which would require a question classification model.

One aspect that is common for both the question and the answer is personalization. We envision the ideal Q\&A system as one that provides questions and answers that are tailored to the specific user. Specifically, we would like to propose questions that are a good fit for the user's background knowledge and interests. To achieve that, one way would be to leverage information about past purchases of the user that can be included in a personalized prompt to generate specific questions and answers for them.

Finally, deploying and maintaining an LLM-based Q\&A recommendation system also poses challenges for high latency and costs associated with the inference of a large model. Some solutions for that can include caching of the questions, offline generation for the head of the traffic, and online ranking of offline generated questions.

\section{Evaluation and Quality Control}
The implementation of mechanisms for quality control of the Q\&A feature are essential for maintaining user trust and ensuring a consistent and helpful experience at production-scale. One of the main challenges for implementing such a mechanism is the large scale of possible input contexts which include the products in the catalog as well as search queries and query prefixes. A reliable quality control mechanism at this scale can be achieved by using a combination of human annotations, automatic (LLM-based) evaluation, and online engagement signals. 
The quality control should verify two main aspects. First, the questions should satisfy some minimal requirements to be acceptable. Second, the questions should be useful to the users, where there can several degrees of usefulness. 

The minimal requirements for questions should be assessed in a point-wise manner by examining each question independently with either a human annotator or an LLM. Below are some of the basic requirements that should be considered.
\begin{itemize}
    \item Relevance: At the minimum, the questions presented should be relevant to the context in which they are presented. For answers, we would like to ensure that they are at least relevant to the questions.
    \item Hallucination-free: It is important to verify that the number of hallucinating cases is minimal and under a predefined threshold. This is crucial for maintaining user trust since they are relying on the provided information to make a purchase decision.
    \item Style: The questions should be natural sounding and well-formed. Furthermore, depending on other user experience guidelines, we would like to ensure that the questions and answers are following a specific style. For example, showing concise questions in friendly tone can be such a requirement.  
    \item Trust and Safety: At the minimum, we must ensure that we do not generate questions and answers that contain information that can hurt the trust of the users, be offensive, or result in actions that are not safe. 
\end{itemize}

For testing the usefulness of questions, the main goal is to find to what extent the questions and answers can help users to make a purchase decision given the context in which they are shown. The evaluation of usefulness can be achieved by leveraging three complementary approaches. First, using online engagement signals such as clicks or purchases can provide indication for the usefulness of questions. While this approach directly measures the customer engagement, it can contain noise and thus can only focus on the head of the traffic. Furthermore, it can be challenging to factor out other reasons for the user actions. Another approach would be to rely on LLM annotations. This approach can be done at scale to increase the testing coverage, but can biased to the LLM assumptions of what is considered useful. Finally, human annotations can be used at a smaller scale, but the utility of a question is subjective and can depend on the knowledge of the annotator, thus requiring multiple annotators to achieve reliability.

Building a representative evaluation data set is another challenge that needs to be addressed at production scale. The data set should be constructed carefully to truthfully represent the product catalog and the query log. Some factors to consider when sampling the data are the product categories, query specificity, product popularity, and considering both the head and the tail of the query log distribution. Ideally, we would want to create two evaluation data sets. The first one would be a static one which would not change over time and is useful for validating model and system changes. The second one would be a dynamic data set which is changing from time to time and can help assess the influence of seasonal trends, such as holidays and special events.

\section{Optimizing Engagement with User Feedback Loops}
\label{sec:opt}
How to optimize the Q\&A recommendation  system with LLMs is an important aspect of the system. While this topic has not been investigated extensively due the recency of LLM-based recommender systems, it is likely to attract growing interest. For example, \citet{senel2024generativeexploreexploittrainingfreeoptimization} recently investigated how generative explore-exploit methods could be used to optimize cohort-level user engagement with LLM-generated content, as measured by click through rates (CTR). There are several challenges that need to be addressed for the effective optimization of the Q\&A recommendation problem. 

First, correctly leveraging users engagement signals is a key factor for the success of the optimization process. Engagement signals can include various direct user actions such as question clicks and hovering over the Q\&A widget. Other actions are more indirect and can occur later in the session such as product purchase or add-to-cart. While these actions can indicate helpfulness of previous interactions with the Q\&A widget, they can also be potentially attributed to other types of interactions. Ideally, the optimization problem should correctly balance between the different types of engagement signals according to the shopping stage of the user and the question type. For example, clicking on informational questions at an early stage can indicate very strong engagement even if there is no purchase at the end of the session, since the customer is still learning about the product space. This can be achieved through techniques such as multi-objective ranking \citep{carmel2020multi}, feature representation, and implementing specific optimization mechanisms for the different locations of the Q\&A widget (e.g., SERP vs. detail page).

Another aspect that is needed to be addressed is the specific optimization methodology. One idea can be to use traditional online approaches like multi-armed bandit \citep{slivkins2019introduction} or learning-to-rank \citep{liu2009learning} to select the best questions. 
A major disadvantage of those approaches is that they cannot directly influence the question and answer generation by the LLM. This may slow down the question improvement process and is limited in terms of exploration of new questions. To try to mitigate this problem, another idea for the optimization would be to guide the LLM in generating questions to optimize user engagement. For example, for a given context we can provide examples of questions that resulted in high click-through rate and prompt the model to generate some new ones. Other more established techniques like RLHF \citep{christiano2017deep,ouyang2022training} and DPO \citep{rafailov2024direct} can be used to this end. For example, we can use engagement data to create pairwise preferences for optimization. Finally, one can design a hybrid approach to leverage both the traditional algorithms as well as LLMs. For example, selecting the important question features based on traditional online algorithms and providing them to the LLM to guide the generation can be one approach in this direction.

Due to the large scale of the Q\&A system, all optimization approaches can suffer from data sparsity problems. One implication of this is that it may take a significant amount of time to obtain sufficient engagement signals that are representative for a large number of contexts. One idea can be to apply optimization first on just the head of the traffic and extend it gradually. Another approach that can be studied would be to use an LLM as a simulated user to generate synthetic data for the optimization.

\section{Conclusion}
In this opinion paper we described our vision for the Q\&A recommendation problem to improve the experience of users in online shopping. The recent success of LLMs has opened up the opportunity to address this problem and bridge the gap between information seeking and product search systems. Implementing and maintaining such an LLM-based system at production scale poses different challenges that need to be addressed and require research and engineering innovations. Those challenges include the controllability of the Q\&A generation, ranking optimization, quality control, and latency. The paper outlined different ideas to address those problems to encourage the research on the topic.

\end{document}